\documentclass{article} 
\usepackage[latin1]{inputenc}
\usepackage{mathptmx}
\usepackage{helvet}
\usepackage{courier}
\usepackage{makeidx}
\usepackage{graphicx}
\usepackage{multicol}
\usepackage{color}
\usepackage{multirow}
\usepackage{url}

\makeindex             

\author{Thomas Philip Runarsson\thanks{University of Iceland, email {\sf tpr@hi.is}} \and Juan J. Merelo-Guervós \thanks{Department of Architecture and Computer
Technology, ETSIIT, University of Granada, Spain, email {\sf jmerelo@geneura.ugr.es}  }}

\title{Adapting Heuristic Mastermind Strategies to Evolutionary Algorithms}

\begin{document}
\maketitle
\begin{abstract}
The art of solving the Mastermind puzzle was initiated by
Donald Knuth and is already more than 30 years old; despite that, it
still receives much attention in operational research and computer
games journals, not to mention the nature-inspired stochastic algorithm
literature. In this paper we try to suggest a strategy that will allow
nature-inspired algorithms to obtain results as good as those based on
exhaustive search strategies; in order to do that, we first review,
compare and improve current approaches to solving the puzzle; then we
test one of these strategies with an estimation of distribution algorithm.
Finally, we try to find a strategy that falls short of being
exhaustive, and is then amenable for inclusion in nature inspired
algorithms (such as evolutionary of particle swarm algorithms). This
paper proves that by the incorporation of local entropy into the fitness function
of the evolutionary algorithm it becomes a better player than a random one,
and gives a rule of thumb on how to incorporate the best heuristic
strategies to evolutionary algorithms without incurring in an excessive
computational cost.
\end{abstract}


\section{Introduction}

Mastermind in its current version is a board game that was
introduced by the telecommunications expert Mordecai Merowitz
\cite{wiki:mm} and sold to the company Invicta Plastics, who renamed
it to its actual name; in fact, Mastermind is a version of a
traditional puzzle called {\em bulls and cows} that dates back to the
Middle Ages. In any case, Mastermind is a puzzle (rather than a game)
in which two persons, the {\em codemaker} and {\em codebreaker} try to
outsmart each other in the following way:

\begin{itemize}
\item The codemaker sets a length $\ell$ combination of $\kappa$
  symbols. In the classical version, $\ell=4$ and $\kappa=6$, and
  color pegs are used as symbols over a board with rows of $\ell=4$
  holes; however, in this paper we will use uppercase letters starting
  with A instead of colours.
\item The codebreaker then tries to guess this secret code by
  producing a combination.
\item The codemaker gives a response consisting on the number of
  symbols guessed in the right position (usually represented as black
  pegs) and the number of symbols in an incorrect position(usually
  represented as white pegs).
\item The codebreaker then, using that information as a hint, produces
  a new combination until the secret code is found.
\end{itemize}

For instance, a game could go like this: The codemaker sets the
secret code {\em ABBC}. The rest of the game is shown in Table
\ref{tab:mm}.

\begin{table}[htbp]
\centering
\begin{tabular}{|l|c|}
\hline
\emph{Combination} & \emph{Response} \\
\hline
AABB & 2 black, 1 white\\
ACDE & 1 black, 1 white\\
FFDA & 1 white\\
ABBE & 3 black \\ 
ABBC & 4 black\\
\hline
\end{tabular}
\vspace{5mm}
\caption{Progress in a Mastermind game that tries to guess the secret
  combination {\em ABBC}. The player here is not particularly clueful,
  playing a third combination that is not {\em consistent} with the
  first one, not coinciding in two positions and one color
  (corresponding to the 2 black/1 white response given by the codemaker)
  with it. \label{tab:mm}}

\end{table}

Different variations of the game include giving information on which
position has been guessed correctly, avoiding repeated symbols in the
secret combination ({\em bulls and cows} is actually this way), or
allowing the codemaker to change the code during the game (but only if
this does not make responses made so far false).

In any case, the codebreaker is allowed to make a maximum number of
combinations (usually fifteen, or more for larger values of $\kappa$
and $\ell$), and score corresponds to the number of combinations
needed to find the secret code; after repeating the game a number of
times with codemaker and codebreaker changing sides, the one with the
lower score wins.

Since Mastermind is asymmetric, in the sense that the position of one
of the players after setting the secret code is almost completely
passive, and limited to give hints as a response to the guesses of the
codebreaker, it is rather a puzzle than a game, since the
codebreaker is not really matching his skills against the codemaker,
but facing a problem that must be solved with the help of hints, the
implication being that playing Mastermind is more similar to solving a
Sudoku than to a game of chess; thus, the solution
to Mastermind, unless in a very particular situation (always playing
with an opponent who has a particular bias for choosing codes, or
maybe playing the dynamic code version), is a search problem with
constraints.

What makes this problem interesting is its relation to other, 
generally called {\em oracle} problems such as circuit and program
testing, differential cryptanalysis and other puzzle games (these
similarities were reviewed in our previous paper \cite{mastermind05})
is the fact that it has been proved to be NP-complete
\cite{abs-cs-0512049, Kendall200813} and that there are several open
issues, namely, what is the lowest average number of guesses you can
achieve, how to minimize the number of evaluations needed to find
them (and thus the run-time of the algorithm), and obviously, how it
scales when increasing $\kappa$ and $\ell$. This paper will concentrate
on the first issue. 

This NP completeness implies that it is difficult to find algorithms
that solve the problem in a reasonable amount of time, and that is why
in our previous work \cite{mastermind05,genmm99,jj-ppsn96} we
introduced  stochastic evolutionary  and simulated annealing  algorithms
that solved the
Mastermind puzzle in the general case, finding solutions in a
reasonable amount of time that scaled roughly logarithmically with
problem size. The strategy followed to play the game was optimal in
the sense that is was guaranteed to find a solution after a finite
number of combinations; however, there was no additional selection on
the combination played other than the fact that it was consistent with
the responses given so far.

In this paper, after reviewing how the state of the art in solving
this puzzle has evolved in the last few years, we examine how we could
improve the code-breaking skills of an evolutionary algorithm by using
different techniques, and how these techniques can be further
optimized. In order to do that we examine different ways of scoring
combinations in the search space, how to choose one combination out of
a set of combinations that have exactly the same score, and how all
that can be applied to a simple estimation of distribution
algorithm to improve results over a standard one. This paper presents
for the first time an evolutionary algorithm that biases search so that
combinations played have a better chance of reducing the size of the
remaining search space, and adapt to an stochastic environment
deterministic techniques that had been previously published; all
techniques, unlike our former papers, have been tested over the whole
code space, instead of a random sample, so that they can be compared
and yield significant results.


The rest of the paper is organized as follows: next we establish
terminology and examine the state of the art; then heuristic
strategies for Mastermind are examined in Section \ref{sec:heu}; the
way they could be adapted to an evolutionary algorithm is presented in
Section \ref{sec:ea}, and finally, conclusions are drawn in the
closing section \ref{sec:fin}

\section{State of the art}
\label{sec:soa}

Before presenting the state of the art, a few definitions are
needed. We will use the term {\em response} for the return code of the
codemaker to a played combination, $c_{played}$. A response is
therefore a function of the combination, $c_{played}$ and the secret
combination $c_{secret}$, let the response be denoted by
$h(c_{played},c_{secret})$.  A combination $c$ is {\em consistent
  with} $c_{played}$ iff
\begin{equation}
h(c_{played},c_{secret}) = h(c_{played},c)
\end{equation}
that is, if the combination has as many black and white pins with
respect to the played combination as the played combination with
respect to the secret combination. Furthermore, a combination is
\emph{consistent} iff
\begin{equation}
h(c_i, c) = h(c_i, c_{secret}) \mbox{ for } i=1..n
\end{equation}
where $n$ is the number of combinations, $c_i$, played so far; that
is, $c$ is {\em consistent with} all guesses made so far. A
combination that is consistent is a candidate solution. The concept of
consistent combination will be important for characterizing different
approaches to the game of Mastermind.

One of the earliest strategies, by Knuth \cite{Knuth}, is perhaps the
most intuitive for Mastermind. In this strategy the player selects the
guess that reduces the number of remaining consistent guesses and the
opponent the return code leading to the maximum number of
guesses. Using a complete minimax search Knuth shows that a maximum of 5
guesses are needed to solve the game using this strategy. This type of strategy is still the most widely used today: most
algorithms for Mastermind start by searching for a 
consistent combination to play.  

In some cases once a single consistent guess is found it is immediately played, in which case the
object is to find a consistent guess as fast as possible. For example,
in \cite{mastermind05} an evolutionary algorithm is described for this
purpose. These strategies are fast and do not need to examine a big
part of the space. Playing a consistent combinations eventually
produces a number of guesses that uniquely determine  the
code. However, the maximum, and average, number of combinations needed
is usually high. Hence, some bias must be introduced in the way
combinations are searched. If not, the guesses will be no better than
a purely random approach, as solutions found (and played) are a random
sample of the space of consistent guesses.

The alternative to discovering a single consistent guess is to collect
a set of consistent guesses and select among them the best
alternative. For this a number of heuristics have been developed over
the years. Typically these heuristics require all consistent guesses
to be first found.  The algorithms then use some kind of search over
the space of consistent combinations, so that only the guess that
extracts the most information from the secret code is issued, or else
the one that reduces as much as possible the set of remaining
consistent combinations. However, this is obviously not known in
advance. To each combination corresponds a partition of the rest of
the space, according to their match (the number of blacks and white
pegs that would be the response when matched with each other). Let us
consider the first combination: if the combination considered is AABB,
there will be 256 combinations whose response will be 0b, 0w (those
with other colors), 256 with 0b, 1w (those with either an A or a B),
etc. Some partitions may also be empty, or contain a single element
(4b, 0w will contain just AABB, obviously).  For a more exhaustive
explanation see \cite{Kooi200513}. Each combination is thus
characterized by the features of these partitions: the number of
non-empty ones, the average number of combinations in them, the
maximum, and other characteristics one may think of.

The path leading to the most successful strategies to date include using
the \emph{worst case}, \emph{expected case},
\emph{entropy} \cite{Neuwirth,bestavros} and \emph{most parts}
\cite{Kooi200513} strategies. The \emph{entropy} strategy selects the
guess with the highest entropy. The entropy is computed as follows:
for each possible response $i$ for a particular consistent guess, the
number of remaining consistent guesses is found. The ratio of
reduction in the number of guesses is also the {\em a priori} probability, $p_i$, of
the secret code being in the corresponding partition. The entropy is
then computed as $\sum_{i=1}^np_i\log_2(1/p_i)$, where $\log_2(1/p_i)$
is the information in bit(s) per partition, and can be used to select
the next combination to play in Mastermind \cite{Neuwirth}. The \emph{worst case} is a one-ply version of Knuth's approach, but Irving \cite{Irving} suggested using the \emph{expected case} rather than the worst case.  Kooi
\cite{Kooi200513} noted, however, that the size of the partitions is
irrelevant and that rather the number of non empty partitions created,
$n$, was important. This strategy is called \emph{most parts}. The
strategies above require one-ply look-ahead and either determining the
size of resulting partitions and/or the number of them. Computing the
number of them is, however, faster than determining their size. For
this reason the \emph{most parts} strategy has a computational
advantage.

The heuristic strategies described above use some form of look-ahead
which is computationally expensive. If no look-ahead is used to guide
the search a guess is selected purely at \emph{random}. However, it
may be possible to discriminate by using local information. If this were possible
one could even dismiss searching for all consistent guesses and
search for a single consistent guess with the bias. In section~\ref{sec:heu}
these heuristic strategies are compared. In section~\ref{sec:EDA} an EDA using
only local information is compared with those that need to examine all
consistent guessed in order to select the best one.

\section{Comparison of heuristic strategies}
\label{sec:heu}

As has been mentioned before, there have been a number of different
strategies proposed over the years for selecting among consistent
guesses in Mastermind. These heuristics do not consider an exhaustive
minimax search, but rather one-ply search. What is, however, not clear
in these research papers is how ties are broken, which probably
implies that a {\em first come, first served} approach is taken, using
the first combination in lexicographical order out of all tied
combinations. For 
this reason we propose to perform a comparison of the heuristic
methods here where the ties are broken randomly. Each strategy is,
therefore, used on all possible secret combinations (they are
$6^4=1296$) using ten independent runs.  

The heuristics compared are the \emph{entropy}, \emph{most parts} and
\emph{worst case} strategy, as performed by Bestavros and Belal
\cite{bestavros}.  The worst case refers to the fact that for each
possible return code for a particular guess the smallest reduction in
assumed, i.e. the worst case. The actual consistent guess chosen is
the one which maximizes the worst case.  Finally, the \emph{expected
  size} strategy, \cite{Irving} is also tested; in this strategy the
expected case is used instead of the worst case. These strategies are
compared with the \emph{random} strategy.

\begin{table}[htbp!]
\centering
\begin{tabular}{|l|rrrrr|c|}
\hline
\emph{Strategy} & \emph{min} & \emph{mean} & \emph{median} & \emph{max}  & \emph{st.dev.} & \emph{max} \\
& & & & & &  \emph{guesses}\\ \hline\hline
Entropy & 4.383 & 4.408 & 4.408 & 4.424 & 0.012 & 6\\
Most parts & 4.383 & 4.410 & 4.412 & 4.430 & 0.013 & 7\\ \hline
Expected size & 4.447 & 4.470 & 4.468 & 4.490 & 0.015 & 7\\
Worst case & 4.461 & 4.479 & 4.473 & 4.506 & 0.016 & 6 \\ \hline
Random & 4.566 & 4.608 & 4.608 & 4.646 & 0.026 & 8\\
\hline
\end{tabular} 
\caption{A comparison of the mean number of games played using all
  $6^4$ colour combinations and breaking ties randomly, ranked from
  best to worst average number of guesses needed. Statistics are
  given for $10$ independent experiments. The maximum number of moves
  used for the $10\times 6^4$ games is also presented in the final
  column. Horizontal separators are given for statistically
  independent results.}\label{tbl:heuristicomparison}
\end{table}

The results of the experiments are given in
table~\ref{tbl:heuristicomparison}. The first combination played is
always AABC, as proposed by \cite{Irving}. The Wilcoxon rank sum (used
instead of t-test since the variable does not follow a normal
distribution) with a 0.05 significance level is
used to determine which results are statistically different form
another. The horizontal lines are used to group together heuristics
that are not statistically different from the other. From these
results we can gather that there is no statistical difference between
the \emph{entropy} and \emph{most parts} strategies.  However, out of
all games played the maximum number of guesses needed by the Entropy
strategy was only 6 while for most parts it was 7. These strategies
are also better than the \emph{worst} and \emph{expected} case, which
are statistically equivalent. For the worst case strategy used,
nevertheless, only a maximum of 6 guesses, unlike the expected case
with 7.  The worst performer is the \emph{random} strategy which also
required a maximum of 8 guesses. Finally, note that the optimal expected result 
on playing all secrets is $4.340$ \cite{Koyama}.

\section{Estimation of distribution algorithm using local entropy}
\label{sec:EDA}

The common approach to using evolutionary algorithms for Mastermind,
is simply to search for a single consistent guess which is then
immediately playing it. This is especially true for the generalized
version of the game, for $N>6$ and $L>4$, where the task of just
finding a consistent guess can be difficult. The result of such an
approach is likely to do as well as the random strategy discussed in
the previous sections. For steady state evolutionary algorithms it may,
however, be the case that the consecutive consistent guesses found may
be similar to others played before. That is, the strategy of play may
not necessarily be purely random. In any case it is highly likely that
evolutionary algorithms of this type will not do better than the
random strategy, as seen above, since consistent combinations found are
a random sample of the set of consistent combinations.

In this section we investigate the performance of strategies that find
a single consistent guess and play it immediately. In this case we use an estimation of
distribution algorithm \cite{eda} {\em EDA} included with the {\tt
  Algorithm::Evolutionary} Perl module \cite{ae09}, with the whole
  EDA-solving algorithm 
available as {\tt Algorithm::MasterMind::EDA} from CPAN (the
comprehensive Perl Archive Network). This is an standard EDA that uses
a population of 200 individuals and a replacement rate of 0.5; each
generation, half the population is generated from the previously
generated distribution. The first combination played was AABB, since it
was not found significantly different from using AABC, as before.

The fitness function used previously \cite{mastermind05} to find
consistent guesses is as follows, 
$$f(c_{guess}) = \sum_{i=1}^n |h(c_i, c_{guess})-h(c_i,c_{secret})|$$
that is, the sum of the absolute difference of the number of white and
black pegs needed to make the guess consistent. However, this approach
is likely to perform as well as the random strategy discussed in the
previous section. When finding a single consistent guess we cannot
apply the heuristic strategies from the previous section. For this
reason we introduce now a local entropy measure, which can be applied
to non-consistent guesses and so bias our search. The local entropy
assumes that the fact that some combinations are better than others
depends on its informational content, and that in turn depends on the
entropy of the combination along with the rest of the combinations
played so far. To compute {\em local entropy}, the combination is
concatenated with $n$ combinations played so far and its Shannon
entropy computed:
\begin{equation}
s(c_{guess}) = \sum_{g \in \{A,...,F\}} \frac{\# g}{(n+1)\ell} \log \bigg( \frac{(n+1)\ell}{\#g}\bigg) 
\end{equation}
with $g$ being a symbol in the alphabet and $\#$ denotes the number of them. 
Thus, the fitness function which includes the local entropy is defined as,
$$f_\ell(c_{guess}) = \frac{s(c_{guess})}{1+f(c_{guess})}$$
In this way a bias is introduced to the fitness to as to select the
guess with the highest local entropy.  When a consistent
combination is found, the combination with the highest entropy found
in the generation is played (which might be the only one or one among
several; however, no special provision is done to generate several).

The result of ten independent runs of the EDA over the whole search
space are now compared with
the results of the previous section.  These results may be seen in
table~\ref{tbl:strategycomparison}. Two EDA experiments are shown, one
using the fitness function designed to find a consistent guess only
($f$) and ones using local entropy $f_\ell$. The EDA using local
entropy is statistically better than playing pure random, whereas the
other EDA is not. In order to confirm the usefulness of the local
entropy, an additional experiment was performed. This time, as in the
previous sections, all consistent guesses are found and the one with
the highest local entropy played. This results is labelled
\emph{LocalEntropy} in table~\ref{tbl:strategycomparison}. The results are
not statistically different from the EDA results using fitness function $f_\ell$.

\begin{table}[htbp!]
\centering
\begin{tabular}{|l|rrrrr|c|}
\hline
\emph{Strategy} & \emph{min} & \emph{mean} & \emph{median} & \emph{max}  & \emph{st.dev.} & \emph{max} \\
& & & & & &  \emph{guesses}\\ \hline\hline
Entropy & 4.383 & 4.408 & 4.408 & 4.424 & 0.012 & 6\\
Most parts & 4.383 & 4.410 & 4.412 & 4.430 & 0.013 & 7\\ \hline
Expected size & 4.447 & 4.470 & 4.468 & 4.490 & 0.015 & 7\\
Worst case & 4.461 & 4.479 & 4.473 & 4.506 & 0.016 & 6 \\ \hline
LocalEntropy & 4.529 & 4.569 & 4.568 & 4.613 & 0.021 & 7\\
EDA+$f_\ell$ & 4.524 & 4.571 & 4.580 & 4.600 & 0.026 &  7 \\ \hline
EDA+$f$ & 4.562 & 4.616 & 4.619 & 4.665 & 0.032 & 7 \\
Random & 4.566 & 4.608 & 4.608 & 4.646 & 0.026 & 8\\
\hline

\end{tabular} 
\caption{A comparison of the mean number of games played using all
  $6^4$ colour combinations and breaking ties randomly, ranked from
  best to worse mean number of combinations. Statistics are
  given for $10$ independent experiments. The maximum number of moves
  used for the $10\times 6^4$ games is also presented in the final
  column. Horizontal separators are given for statistically
  independent results. }\label{tbl:strategycomparison} 
\end{table}

As a local conclusion, the {\em Entropy} method seemed to perform the
best on average, but the estimation of distribution algorithm is not
statistically different from (admittedly naive) exhaustive search
strategies such as LocalEntropy and performs
significantly better than the Random algorithm on average. 

We should
remark that the objective of this paper is not to show which strategy is the
best runtime-wise, or which one offers the best algorithmic
performance/runtime trade-off; but in any case we should note that the
algorithm with the least number of evaluations and lowest runtime is
the EDA. However, its average performance as a player is not as good as the rest,
so some improvement might be obtained by creating a set of possible
solutions. It remains to be seen how many solutions would be needed,
but that will be investigated in the next section.

\section{Heuristics based on a subset of consistent guesses}
\label{sec:ea}

Following a tip in one of our former papers, recently Berghman et al.
\cite{Berghman20091880} 
proposed an evolutionary algorithm which finds a number of consistent
guesses and then uses a strategy to select which one of these should
be played. The strategy they apply is not unlike the \emph{expected
  size} strategy. However, it differs in some fundamental ways. In
their approach each consistent guess is assumed to be the secret in
turn and each guess played against every different secret. The return
codes are then used to compute the size of the set of remaining consistent
guesses in the set. An average is then taken over the size of these
sets. Here, the key difference between the 
\emph{expected size} method is that only a subset of all possible
consistent guesses is used and some return codes may not be considered
or considered more frequently than once, which might lead to a bias in
the result. Indeed they remark that their approach is
computationally intensive which leads them to reduce the size of this subset further.
Note that Berghman et al. only present the result of a single
evolutionary run and so their results cannot be compared with those
here. 

Their approach is, however, interesting, and lead us to consider the
case where an evolutionary algorithms has been designed 
to find a maximum of $\mu$ consistent guesses within some finite
time. It will be assumed that this subset is sampled uniformly and
randomly from all possible consistent guesses. The question is, how do
the heuristic strategies discussed above work on a randomly sampled subset
of consistent guesses? The experiment performed in the previous
sections are now repeated, but this time only using the four best one-ply
look-ahead heuristic strategies on a random subset of guesses, bounded by size
 $\mu$. If there are many guesses that give the same number
of partitions or similar entropy then perhaps taking a random subset
would be a good representation for all guesses. This has implications
not only with respect to the application of EAs  but also to the common
strategies discussed here.   

The size of the subsets are fixed at $10$, $20$, $30$, $40$, and $50$,
in order to investigate the influence of the subset size. The results
for these experiments and their statistics are presented in
table~\ref{tbl:redstrategycomparison}. The results are presented are
as expected better as the subset size, $\mu$, gets bigger. Noticeable
is the fact that the \emph{entropy} and \emph{most parts} strategies perform the
best as before, however, at $\mu=40$ and $50$ the entropy strategy is
better.  

\begin{table}[htbp!]
\centering
\begin{tabular}{|l|rrrrr|c|}
\hline
\emph{Strategy} & \emph{min} & \emph{mean} & \emph{median} & \emph{max}  & \emph{st.dev.} & \emph{max} \\
& & & & & &  \emph{guesses}\\ \hline\hline
$\mu=10$ & & & & &  & \\
Most parts & 4.429 & 4.454 & 4.454 & 4.477 & 0.016 & 7 \\\hline
Entropy & 4.438 & 4.468 & 4.476 & 4.483 & 0.016 & 7\\
Expected size & 4.450 & 4.472 & 4.474 & 4.493 & 0.014 & 7\\\hline
Worst case & 4.447 & 4.486 & 4.487 & 4.519 & 0.020 & 7\\ \hline\hline
$\mu=20$ & & & & & & \\
Entropy & 4.394 & 4.423 & 4.426 & 4.455 & 0.021 & 7\\
Most parts & 4.424 & 4.431 & 4.427 & 4.451 & 0.009 & 7\\\hline
Expected size & 4.427 & 4.454 & 4.455 & 4.481 & 0.017 & 7\\
Worst case & 4.429 & 4.453 & 4.451 & 4.486 & 0.017 & 7\\\hline\hline
$\mu=30$ & & & & & & \\
Entropy & 4.380 & 4.413 & 4.410 & 4.443 & 0.020 & 6\\
Most parts & 4.393 & 4.416 & 4.416 & 4.435 & 0.015 & 7\\\hline
Expected size & 4.426 & 4.453 & 4.456 & 4.491 & 0.019 & 7 \\
Worst case & 4.434 & 4.459 & 4.461 & 4.477 & 0.013 & 7\\\hline\hline
$\mu=40$ & & & & & & \\
Entropy & 4.372 & 4.398 & 4.399 & 4.426 & 0.017 & 7\\ \hline
Most parts & 4.383 & 4.424 & 4.427 & 4.448 & 0.020 & 7\\ \hline
Expected size & 4.418 & 4.457 & 4.455 & 4.491 & 0.023 & 7\\
Worst case & 4.424 & 4.458 & 4.457 & 4.490 & 0.022 & 7\\\hline\hline
$\mu=50$ & & & & & & \\
Entropy & 4.365 & 4.397 & 4.393 & 4.438 & 0.020 & 6 \\\hline
Most parts & 4.400 & 4.424 & 4.422 & 4.454 & 0.017 & 7\\\hline
Expected size & 4.419 & 4.453 & 4.453 & 4.495 & 0.022 & 7 \\
Worst case & 4.431 & 4.456 & 4.457 & 4.474 & 0.012 & 6\\\hline
\end{tabular} 
\caption{Statistics for the average number of guesses for different maximum sizes $\mu$ of subsets of consistent guesses. The horizontal lines are used as before to indicate statistical independent, with the exception of one case: for $\mu=10$ the expected size and worst case are not independent.}\label{tbl:redstrategycomparison}
\end{table}

Is there a statistical difference between the different subset sizes?
To answer this we look at only the two best strategies in more detail,
\emph{entropy} and \emph{most parts}, and compare their performances 
for the different subset sizes, $\mu$, and using the complete set, case when $\mu=\infty$, as
presented in table~\ref{tbl:strategycomparison}.  These results are
given in table~\ref{tbl:redstrategycomparisonentropy} and
\ref{tbl:redstrategycomparisonmostparts}. From this analysis it may be
concluded that a set size of $\mu=20$ is sufficiently large and not
statistically different from using the entire set of consistent
guesses. This is actually quite a large reduction is the set size,
which is about 250 on average after the first guess, then 55, followed
by 12 \cite{Berghman20091880}.

\begin{table}[h!]
\centering
\begin{tabular}{|l|rrrrr|}
\hline
$\mu=$ & \emph{min} & \emph{mean} & \emph{median} & \emph{max}  & \emph{st.dev.} \\\hline\hline
10 & 4.438 & 4.468 & 4.476 & 4.483 & 0.016 \\ \hline
20 & 4.394 & 4.423 & 4.426 & 4.455 & 0.021 \\\hline
30 & 4.380 & 4.413 & 4.410 & 4.443 & 0.020 \\
40 & 4.372 & 4.398 & 4.399 & 4.426 & 0.017 \\
50 & 4.365 & 4.397 & 4.393 & 4.438 & 0.020 \\
$\infty$ & 4.383 & 4.408 & 4.408 & 4.424 & 0.012 \\\hline
\end{tabular} 
\caption{No statistical advantage is gained when using a set size
  larger than $\mu=30$ when using the \emph{entropy}
  strategy. However, there is also no statistically difference between
  $\mu=20$ and both $\mu=30$ and $\mu=\infty$ (the only cases not
  indicated by the horizontal lines).}\label{tbl:redstrategycomparisonentropy} 
\end{table}

\begin{table}[h!]
\centering
\begin{tabular}{|l|rrrrr|}
\hline
$\mu=$ & \emph{min} & \emph{mean} & \emph{median} & \emph{max}  & \emph{st.dev.} \\\hline\hline
10 & 4.429 & 4.454 & 4.454 & 4.477 & 0.016 \\\hline
20 & 4.424 & 4.431 & 4.427 & 4.451 & 0.009 \\
30 & 4.393 & 4.416 & 4.416 & 4.435 & 0.015 \\
40 & 4.383 & 4.424 & 4.427 & 4.448 & 0.020 \\
50 & 4.400 & 4.424 & 4.422 & 4.454 & 0.017 \\
$\infty$ & 4.383 & 4.410 & 4.412 & 4.430 & 0.013 \\\hline
\end{tabular} 
\caption{No statistical advantage is gained when using a set size
  larger than $\mu=20$ for the \emph{most parts} strategy. However,
  there is a statistical difference between $\mu=20$ and $\mu=\infty$
  (the only case not indicated by the horizontal
  lines.}\label{tbl:redstrategycomparisonmostparts} 
\end{table}

This implies that, at least in this case, using a subset of the
combination pool that is around $1/10$th of the total size potentially
yields a result that is as good 
as using the whole set; even as algorithmically finding 20 tentative
solutions is harder than finding a single one, using this in
stochastic search algorithms such as the EDA mentioned above or an
evolutionary algorithm holds the promise of combining the accuracy of
exhaustive search algorithms with the speed of an EDA or an EA. In any
case, for spaces bigger than $\kappa=6, \ell=4$ there is no other
option, and this 1/10 gives at least a rule of thumb. How this
proportion grows with search space size is still an open question.

\section{Discussion and Conclusion}
\label{sec:fin}

In this paper we have tried to study and compare the different
heuristic strategies for the simplest version of Mastermind in order to
come up with a nature-inspired algorithm that is able to beat them in
terms of running time and scalability. The main problem with heuristic
strategies is that they need to have the whole search space in memory;
even the most advanced ones that run over it only once will become
unwieldy as soon as $\ell$ or $\kappa$ increase. However, evolutionary
algorithms have already been proved \cite{mastermind05} to scale much
better, the only problem being that their performance as players is no
better than a random player. 

In this paper, after improving (or maybe just clarifying) heuristic and
deterministic algorithms with an random choice of a combination to
play, we have incorporated the simplest of those strategies to an
estimation of distribution algorithm (the so-called \emph{local entropy}, which
takes into account the amount of {\em surprise} the new combination
implies); results are promising, but still fall short of the best
heuristic strategies, which take into account the partition of search
space created by each combination. That is why we have tried to compute
the subset that would be able to obtain results that are
indistinguishable, in the statistical sense, from those obtained with
the whole set, coming up with a subset whose size is around 10\% of the
whole one, being thus less computational intensive and easily
incorporated into an evolutionary algorithm.

However, how this is incorporated within the evolutionary algorithm
remains to be seen, and will be one of our future lines of work. So far,
distance to consistency and entropy are combined in an aggregative
fitness function; the quality of partitions induced will also have to be
taken into account; however, there are several ways of doing this:
putting consistent solutions in an {\em archive}, in the same fashion
that multiobjective optimization algorithms do, leave them into the
population and take the quality of partitions as another objective, not
to mention the evolutionary parameter issues themselves: population
size, operator rate. Our objective, in this sense, will be not only to
try and minimize the number of average/median games played, but also to
minimize the proportion of the search space examined to find the final
solution.

All the tests and algorithms have been implemented using the Matlab
package, and are available as open source source software with a GPL
licence from the authors.  The evolutionary algorithm and several
mastermind strategies are also available from CPAN; most results and
configuration files needed to compute them are available from the
group's CVS server.

\section*{Acknowledgements}

This paper has been funded in part by the Spanish MICYT projects NoHNES
(Spanish Ministerio de Educaci\'on y Ciencia - TIN2007-68083) and
TIN2008-06491-C04-01 and the 
Junta de Andaluc\'ia P06-TIC-02025 and P07-TIC-03044. The authors are
also very grateful to the traffic jams in Granada, which allowed
limitless moments of discussion and interaction over this problem. 

\bibliographystyle{spmpsci}      
\bibliography{mastermind,geneura}

\end{document}